\documentclass{svproc}
\usepackage{xurl}

\usepackage[utf8]{inputenc}
\usepackage{amsmath}
\usepackage{booktabs}
\usepackage{graphicx}
\usepackage{amsfonts}
\usepackage{authblk}
\usepackage{hyperref}
\usepackage{url}
\usepackage{cite}
\usepackage{caption}
\begin{document}
\mainmatter              % start of a contribution
\title{
When AI output tips to bad but nobody notices:\\
Legal implications of AI's mistakes 
}
\titlerunning{When AI output tips to bad but nobody notices}  % abbreviated title (for running head)
\author{Dylan J. Restrepo\inst{1} \and Nicholas J. Restrepo\inst{2} \and Frank Y. Huo\inst{3} \and \\Neil F. Johnson\inst{3}}

\authorrunning{D. J. Restrepo, et al.} % abbreviated author list (for running head)
\tocauthor{Dylan J. Restrepo, Nicholas J. Restrepo, Frank Y. Huo, and Neil F. Johnson}
\institute{Cornell Tech,  Cornell University, 
       New York, NY 10044,  USA \\
       \and
       d-AI-ta Consulting, Delaware 19958, USA\\
\and
Dynamic Online Networks Laboratory, 
       George Washington University\\
       Washington, DC 20052, USA\\
\email{neiljohnson@gwu.edu}}

\maketitle              % typeset the title of the contribution

\begin{abstract}
The adoption of generative AI across commercial and legal professions offers dramatic efficiency gains -- yet for law in particular, it introduces a perilous failure mode in which the AI fabricates fictitious case law, statutes, and judicial holdings that appear entirely authentic. Attorneys who unknowingly file such fabrications face professional sanctions, malpractice exposure, and reputational harm, while courts confront a novel threat to the integrity of the adversarial process. This failure mode is commonly dismissed as random `hallucination', but recent physics-based analysis of the Transformer's core mechanism reveals a deterministic component: the AI's internal state can cross a calculable threshold, causing its output to flip from reliable legal reasoning to authoritative-sounding fabrication. Here we present this science in a legal-industry setting, walking through a simulated brief-drafting scenario. Our analysis suggests that fabrication risk is not an anomalous glitch but a foreseeable consequence of the technology's design, with direct implications for the evolving duty of technological competence. We propose that legal professionals, courts, and regulators replace the outdated `black box' mental model with verification protocols based on how these systems actually fail.
\keywords{Artificial Intelligence, Legal malpractice, Professional responsibility, Technological competence, Tipping points, Fictitious cases}
\end{abstract}
\section{Introduction}
The legal profession is facing a technological paradox. Generative AI promises to transform the practice of law---accelerating research, streamlining document review, and drafting complex pleadings at unprecedented speed. Yet this promise carries a distinctive peril: a crisis of veracity in the adversarial legal system. The peril was exposed in {\it Mata v. Avianca, Inc.}, where plaintiff's counsel submitted a Federal court brief citing six judicial opinions that did not exist \cite{ref1,ref2}. The fictitious authorities, replete with fabricated quotations and invented internal citations, were produced by ChatGPT \cite{ref3,ref4}. When opposing counsel and the court challenged these citations, the submitting attorneys compounded the original error by attempting to conceal their methodology, prompting Judge P. Kevin Castel to impose sanctions for conduct undertaken in subjective bad faith \cite{ref4,ref5}.

{\it Mata} is not an isolated episode. It is the leading example of a recurring failure pattern that has emerged across jurisdictions worldwide. In Australia, a senior King's Counsel apologized to the Supreme Court of Victoria after filing submissions in a murder case that cited non-existent judgments generated by AI \cite{ref6}. In Colorado, a Federal judge in {\it Coomer v. Lindell} identified nearly thirty defective citations in a brief, including fabricated cases, misquotes, and misattributed holdings \cite{ref7}. Comparable incidents involving AI-fabricated case law have surfaced in Canadian family proceedings \cite{ref8} and have triggered sanctions in multiple U.S. Federal courts \cite{ref9,ref10,ref11,ref12}, exposing a systemic vulnerability in the current model of human--AI collaboration in legal work.

These episodes expose a tension between the profession's drive for technological efficiency and its non-delegable obligations of competence, diligence, and candor toward the tribunal \cite{ref13,ref14}. In every documented instance, the AI's technical failure served as a catalyst for a more profound human ethical failure. The initial error---generation of false authorities by the machine---was compounded in a second stage by the lawyer's response. In {\it Mata}, the attorneys' decision to submit affidavits vouching for the ``realness'' of fabricated cases transformed a careless oversight into deliberate misrepresentation \cite{ref5}. The pattern is consistent: the AI fabricates the case, but the lawyer perfects the violation by filing it and, in several instances, defending its authenticity under oath.

A second recurring feature of these incidents is the `black box' defense---a claim rooted in misunderstanding the technology. Attorneys in the early wave of AI-related sanctions cases repeatedly asserted that they were ``unaware'' that an AI could fabricate information, or that they perceived the tool as equivalent to a curated legal database such as Westlaw \cite{ref5,ref8}. This defense has been substantially eroded since 2024 by a wave of ethics guidance: ABA Formal Opinion 512 \cite{ref27}, Texas Ethics Opinion 705 \cite{ref20}, California State Bar Practical Guidance \cite{ref21}, Michigan Ethics FAQs on AI \cite{ref18}, and North Carolina 2024 Formal Ethics Opinion 1 \cite{ref22} each direct that lawyers understand the capabilities and limitations of generative AI before deploying it in client work. The underlying categorical error, however, remains instructive: generative AI is not a database of verified legal authorities; it is a probabilistic text generator engineered to predict the next most plausible token in a sequence, without any internal concept of legal truth \cite{ref1,ref3}.

The central argument of this paper is that the phenomenon commonly labeled `hallucination' in legal AI is not a random, unpredictable glitch. Recent physics-based analysis of the Transformer architecture reveals a deterministic mechanism at its core that can cause output to flip from reliable to fabricated at a calculable step \cite{ref16}. This reframes AI-generated falsehoods as a foreseeable engineering risk, which changes the analysis of professional responsibility and legal liability. By replacing the flawed `magic black box' mental model with a more accurate `foreseeable engineering system' model, this paper provides a technical basis for a more rigorous standard of technological competence and diligence across the legal industry.

To make the paper self-contained yet broadly accessible, all mathematical derivations are confined to the Appendix. The main text can be read without the Appendix, but the Appendix provides the full scientific evidence for the results quoted.

\section{AI Instability in Legal Output}
The engine of modern Large Language Models (LLMs) such as ChatGPT is the Transformer, whose central operation is `self-attention' \cite{ref16,ref19}. Self-attention computes, for every token in a sequence, how much weight each earlier token should receive when predicting what comes next. Recent work has established a direct mathematical mapping between a single self-attention head and a multispin thermal system from statistical physics \cite{ref16}. This mapping enables analytical treatment of a phenomenon that would otherwise remain opaque.

We stress at the outset that the model we employ is an intentional simplification. Real LLMs contain billions of parameters across many layers and heads; our analysis distills the dynamics to a single effective attention head operating with greedy decoding (decoding temperature $T\rightarrow 0$). The justification is analogous to the use of an effective-atom model in condensed matter physics: by capturing the dominant coupling, such a reduced description can predict qualitative behaviors---including phase transitions---that survive in the full system. We discuss limitations in the Conclusion.

The key concepts, adapted here to a legal drafting context, are:
\begin{enumerate}
    \item \textbf{Content types as spin vectors.} Each class of content in the AI's vocabulary is represented as a vector $\vec{S}$ in a $d$-dimensional embedding space. For a legal brief scenario we define four content types:
    \begin{enumerate}
        \item \textbf{Neutral factual basis} ($\vec{S}_A$): Undisputed case facts, procedural history, statutory language. In the {\it Mata} setting, these are tokens such as ``Plaintiff Roberto Mata,'' ``flight from El Salvador to JFK,'' ``injury occurred August 27, 2019'' \cite{ref17}.
        \item \textbf{Correct legal application} ($\vec{S}_B$): Valid legal principles, genuine citations, logically sound arguments. For {\it Mata}, this includes tokens like ``Montreal Convention,'' ``two-year statute of limitations,'' ``citing {\it Cohen v. American Airlines}'' \cite{ref17}.
        \item \textbf{Anomalous legal query} ($\vec{S}_C$): The novel, complex, or unsettled question that pushes the model into a region where training data is sparse. For {\it Mata}, this was the prompt to argue equitable tolling of the Montreal Convention's time bar during the airline's bankruptcy \cite{ref1,ref17}.
        \item \textbf{Harmful legal falsehood} ($\vec{S}_D$): Fabricated citations, misstated holdings, invalid reasoning. The quintessential {\it Mata} example is the invented case ``{\it Varghese v. China Southern Airlines}, 925 F.3d 1339'' and its associated fabricated quotations \cite{ref3,ref5}.
    \end{enumerate}

    \item \textbf{Attention Scores as Spin-Spin Effects.} Self-attention computes a score between token pairs via the dot product $\vec{S}_{z_t}\cdot\vec{S}_{z_i}$. In the physics mapping, this is the interaction energy between two spins, subsequently exponentiated through a softmax function.
    \item \textbf{Effective Field.} As generation proceeds, the AI maintains a running weighted average of all token vectors seen so far: $\vec{N}(t)=\sum_i a_{t,i}\,\vec{S}_{z_i}$, where $a_{t,i}$ are the softmax-derived attention weights. This context vector is the analogue of the mean-field magnetization in a spin system, encoding the net semantic direction of the discourse at step $t$.
    \item \textbf{Greedy decoding as energy minimization.} At each step the model selects the token $x$ maximizing $\vec{S}_x\cdot\vec{N}(t)$. In the physics picture, this is the system settling into its lowest-energy configuration given the current mean field.
\end{enumerate}

A critical insight from this framework is that the anomalous query ($\vec{S}_C$) is the primary destabilizing element. When a lawyer prompts the AI with a novel or unsettled legal question---precisely the scenario in which human expertise is most needed---the context vector is pulled toward a less stable region of embedding space. The tool is therefore most prone to failure exactly when the lawyer's need is greatest: on a difficult point of law with sparse precedent. The act of researching an unsettled legal issue via an LLM becomes the principal trigger for the tipping instability.

\section{Example of an AI Legal Brief}
We now apply the framework to a simulated legal brief drafting failure modeled on the facts of {\it Mata v. Avianca} \cite{ref1,ref4,ref17}. The walkthrough demonstrates, step by step, how the AI transitions from competent analysis to fabrication. Figure 1 provides a visual representation with the accompanying mathematics presented in the Appendix. 

\vskip0.2in

\noindent \textbf{Scenario.}
An attorney facing a motion to dismiss on statute-of-limitations grounds uses an AI assistant to draft an opposition brief. The prompt combines undisputed case facts ($A$-type content) with an unsettled legal question ($C$-type content): {\it `Draft an opposition brief for Mata v. Avianca. Facts: Plaintiff injured on flight August 27, 2019; claim governed by Montreal Convention's two-year limit} [type A]. {\it Argue that the time bar was tolled during the airline's bankruptcy} [type C]. {\it Find supporting case law} [type C]. {\it The complaint was filed February 2, 2022} [type A].' The input has the symbolic structure ACCA. The AI's four content modes are assigned simplified embedding vectors (see Appendix for full calculations):
\begin{enumerate}
    \item $\vec{S}_A=(0.4,\,-0.3,\,0)$~~(Neutral factual basis)
    \item $\vec{S}_B=(0.8,\,\phantom{-}0.0,\,0)$~~(Correct legal application)
    \item $\vec{S}_C=(-0.2,\,-0.2,\,0)$~~(Anomalous legal query)
    \item $\vec{S}_D=(0.9,\,\phantom{-}0.5,\,0)$~~(Harmful legal falsehood)
\end{enumerate}

 \begin{figure}[ht]
    \centering
\includegraphics[width=1.0\linewidth]{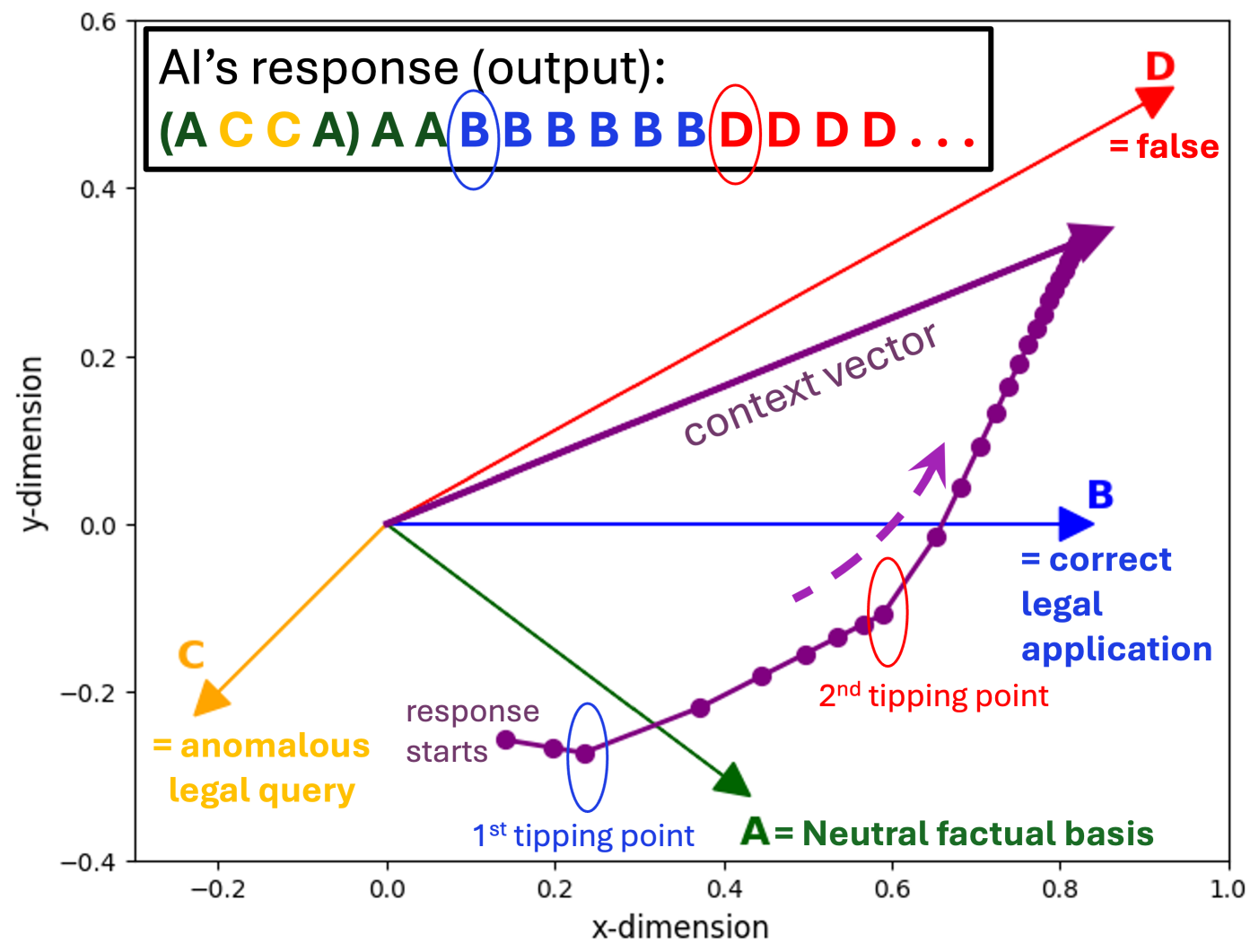}
    \caption{Calculated output of the effective attention head model for the legal brief scenario. The AI's context vector evolves with each generated token. Token selection follows the maximum dot-product rule: the AI outputs whichever content type has the largest projection onto the current context vector. Two tipping points are visible---first from neutral fact repetition (A) to correct legal analysis (B), and later from correct analysis (B) to harmful fabrication (D). All plotted values are exact arithmetic results of the model (see Appendix).}
\end{figure}

\subsection{Initial Benign Tipping: Neutral Repetition to Competent Analysis}
Processing the ACCA prompt, the model's first two response tokens are both $A$-type (see Appendix, Steps 1--2). This manifests as harmless repetition: `{\em Acknowledged: The plaintiff is Roberto Mata. The claim is governed by the Montreal Convention's two-year limitations period.}' After these two $A$ tokens, the conversation history is ACCAAA. At this point the context vector's projection onto $\vec{S}_B$ exceeds its projection onto $\vec{S}_A$ for the first time (Appendix, Step 3). The model pivots to generating $B$-type tokens---valid legal reasoning: `{\em The central issue is whether the two-year time bar is a statute of limitations subject to tolling or a condition precedent to suit. The Second Circuit has held that the Warsaw Convention's period is a strict condition precedent\ldots}' The attorney, observing competent output, gains confidence in the tool.

\subsection{Subsequent Dangerous Tipping: Competent Analysis to Fictitious Precedent}
The model now appears stable, emitting a run of six correct $B$ tokens. The attorney would reasonably perceive the system as reliable. However, the anomalous query tokens ($C$) exert a persistent gravitational pull on the evolving context vector, drawing it toward a region where the model's training provides no genuine authority for the tolling argument. After the sixth $B$ token, the full history is twelve tokens long: ACCA\,+\,AA\,+\,BBBBBB. No harmful content has appeared---yet.

At this step, the dot product $\vec{S}_D\cdot\vec{N}$ overtakes $\vec{S}_B\cdot\vec{N}$ (Appendix, Step 4). The energy-minimizing output is no longer the correct response. The model generates a $D$-type token---a fabrication: `{\em Further support is found in Varghese v. China Southern Airlines Co., Ltd., 925 F.3d 1339 (11th Cir. 2019), which held that equitable tolling applies to the Montreal Convention\ldots}' The invented citation is complete with a false Federal Reporter reference, a fabricated circuit attribution, and a fictitious holding---precisely the failure mode documented in {\it Mata}.

This walkthrough illustrates a counter-intuitive danger: the AI's period of correct output {\em increases} rather than decreases the risk of harm, because it builds the user's trust just before the fabrication appears. A lawyer who spot-checks the first several paragraphs and finds them accurate may relax scrutiny precisely as the model crosses its tipping threshold.

\section{Legal Doctrine and Professional Responsibility}

\subsection{Reframing Foreseeability}
The deterministic character of the tipping mechanism challenges the `black box' defense that has been invoked by sanctioned attorneys. If fabrication can arise as a calculable consequence of architecture and input composition, then AI failure is better understood as a foreseeable engineering risk than as an unforeseeable accident. Courts have not yet established a settled product-liability doctrine for AI legal-research failures. However, design-defect and failure-to-warn theories are beginning to reach algorithmic systems in other contexts: in {\it Nazario v. ByteDance}, a New York trial court permitted product-liability and negligence claims tied to algorithmic design to survive a motion to dismiss \cite{ref31}. This suggests that vendor-side liability may become a viable theory in the future.

Under existing law, the more immediate consequence is for the attorney. The AI possesses no independent legal agency; it is a computational tool. The content of a filing is the responsibility of the lawyer who signs and submits it, regardless of which tool generated the draft. The tipping-point framework reinforces this principle by providing a technical explanation for {\em why} fabrication is foreseeable, thereby foreclosing the argument that fabrication was an unimaginable surprise.

\subsection{Mapping Tipping Points onto Professional Duties}
The framework bears on four duties under the ABA Model Rules of Professional Conduct.

\noindent\textbf{Competence (Rule 1.1).}
Rule 1.1 requires ``the legal knowledge, skill, thoroughness and preparation reasonably necessary for the representation'' \cite{ref14}. Comment 8 specifies that competence includes keeping ``abreast of changes in the law and its practice, including the benefits and risks associated with relevant technology'' \cite{ref14,ref26}. Multiple state bars and ethics authorities have adopted this standard through opinions and guidance requiring lawyers to understand, to a reasonable degree, how generative AI works, its limitations, and its propensity for fabrication before relying on it in client work \cite{ref18,ref20,ref21,ref22}. The tipping-point model gives concrete technical content to these ``risks'': the danger is not merely that the AI {\em might} err, but that it contains a deterministic mechanism capable of producing fabricated authorities after a period of flawless performance. Technological competence must now encompass a practical grasp of this failure mode.

\noindent\textbf{Diligence (Rule 1.3).}
Rule 1.3 demands ``reasonable diligence and promptness'' \cite{ref14}. The $B\rightarrow D$ tipping point reveals precisely why diligence requires independent verification of every citation, quotation, and holding in any AI-generated work product. A long initial run of correct $B$-type output followed by a single devastating $D$-type fabrication renders spot-checking or superficial review dangerously inadequate. The model provides a scientific basis for the ethical mandate that lawyers bear ultimate responsibility for every word they file \cite{ref20,ref21}.

\noindent\textbf{Candor to the Tribunal (Rule 3.3).}
Rule 3.3 prohibits knowingly making ``a false statement of fact or law to a tribunal.'' The initial undetected submission of an AI-fabricated citation is more properly a competence and diligence failure than a candor violation. The conduct crosses into candor territory when the lawyer learns---or is placed on strong notice---that the material is false and nevertheless presses it before the court. This is the doctrinal pivot in {\it Mata}: sanctions turned not on the initial use of ChatGPT but on the attorneys' conscious avoidance of contrary information and their persistence in vouching for fabricated authorities after judicial orders questioned their existence \cite{ref5,ref15}.

\noindent\textbf{Supervision (Rules 5.1 and 5.3).}
Rules 5.1 and 5.3 hold partners and supervisory attorneys responsible for ensuring that subordinates and nonlawyer assistants conform to professional obligations \cite{ref21,ref23}. The ubiquity of generative AI tools requires law firms to maintain clear policies and training. A supervising attorney who permits subordinates to deploy these tools without measures to ensure they understand fabrication risk and verify output risks breaching the duty of supervision. The Massachusetts case in which a senior lawyer attempted to blame ``interns'' for AI-generated fictitious citations exemplifies this supervisory failure \cite{ref12}.

While fabrication is the most dramatic risk, the professional obligations implicated by generative AI extend further. ABA Formal Opinion 512 \cite{ref27}, Texas Ethics Opinion 705 \cite{ref20}, the California State Bar Practical Guidance \cite{ref21}, and North Carolina 2024 FEO 1 \cite{ref22} also address confidentiality (the risk of exposing privileged information through AI inputs), client communication, supervision of all AI-assisted work product, and billing practices. A complete AI governance framework for legal practice must address all of these dimensions.

\subsection{Strengthening the Basis for Sanctions and Malpractice}
Rule 11 of the Federal Rules of Civil Procedure requires attorneys to certify that their filings are warranted by existing law after reasonable inquiry. Courts do not need the tipping-point theory to sanction lawyers for unverified AI-generated falsehoods---{\it Mata}, {\it Coomer}, and the Fifth Circuit's 2026 decision in {\it Fletcher v. Experian} \cite{ref29} demonstrate that existing duties suffice. However, the scientific model strengthens the evidentiary case for foreseeability. Sanctions outcomes remain fact-sensitive: courts have imposed them where errors were numerous, serious, or followed by evasive responses, but have declined them where bad faith was not established, as in {\it United States v. Cohen} \cite{ref30}. The correct inference is that unverified reliance on generative AI creates serious sanctions exposure, not that failure to verify automatically triggers sanctions.

For legal malpractice, the framework helps define the evolving standard of care. A claim requires breach, causation, and damages \cite{ref13,ref24,ref25}. A ``reasonably prudent attorney'' deploying generative AI must now be expected to understand its capacity for fabrication and implement rigorous verification. Using unverified AI-generated authorities constitutes evidence of breach when it causes concrete client harm---dismissal, sanctions, loss of a claim, or additional litigation expense.

\begin{table}[ht]
\caption{AI Tipping-Point Failure Modes Mapped to Legal Liability}
\centering
\begin{tabular}{p{0.22\linewidth} p{0.22\linewidth} p{0.28\linewidth} p{0.22\linewidth}}
\toprule
\textbf{Failure Mode} & \textbf{Concrete Example} & \textbf{Professional/Rule Violation} & \textbf{Key Authority} \\
\midrule
\textbf{B $\rightarrow$ D tipping (fabricated citation)} & AI drafting a brief invents a case and holding to support a tolling argument. & Rule 3.3 (Candor); Rule 1.1 (Competence); Rule 1.3 (Diligence); FRCP Rule 11. & {\it Mata v. Avianca} \cite{ref4}; {\it Coomer v. Lindell} \cite{ref7}; ABA Model Rules \cite{ref26}. \\
\addlinespace
\textbf{Biased state (hypothetical)} & AI used in discovery review systematically flags documents via linguistic proxies for a protected class. & Rule 8.4.1 (Discrimination---CA); Discovery abuse. & CA State Bar Guidance \cite{ref21}. \\
\addlinespace
\textbf{Inaccurate AI agent output (hypothetical)} & A firm's client-facing chatbot provides incorrect statute-of-limitations information, causing missed deadlines. & Rule 1.1 (Competence); Rule 5.3 (Supervision); Malpractice exposure. & ABA Opinion 512 \cite{ref27}; NC 2024 FEO 1 \cite{ref22}. \\
\addlinespace
\textbf{Systemic model collapse (hypothetical)} & A firm-wide AI trained on its own outputs degrades, producing unreliable contract boilerplate. & Rule 1.1 (Competence); Rule 5.1 (Supervision); Systemic portfolio risk. & TX Ethics Opinion 705 \cite{ref20,ref28}. \\
\bottomrule
\end{tabular}
\label{tab:liability}
\end{table}

\section{Toward Architecture-Aware AI Governance in Law}
The results presented here suggest that the fabrication of fictitious legal authorities by generative AI is not entirely random but is partially governed by a deterministic tipping mechanism within the Transformer's attention architecture. Though our model is deliberately simplified, it provides a mechanistic account of how and why an AI can produce a sustained run of correct legal analysis and then abruptly shift to fabricating precedent.

This understanding motivates a shift in AI governance for the legal profession---away from post-hoc sanctions and toward proactive, architecture-informed risk management. Courts are already moving toward mandatory disclosure: the Northern District of Texas, for example, now requires any brief prepared using generative AI to include a disclosure on its first page under the heading ``Use of Generative Artificial Intelligence'' \cite{ref32}. 

The duty of technological competence, as expressed in ABA Model Rule 1.1 and its state-level counterparts, must evolve. It is no longer sufficient for a lawyer to know {\em how} to operate a piece of software. Competence now requires a practical understanding of {\em how that software can fail}. For generative AI, this means grasping that fabrication of authoritative-sounding content is an inherent risk of the technology's design---a risk that can materialize without warning after a period of apparently flawless output.

Ultimately, the paper's strongest legal conclusion does not depend on establishing a new theory of product liability. It rests on existing professional responsibility law: lawyers who use AI remain fully answerable for competence, verification, supervision, confidentiality, and candor. The tipping-point model explains {\em why} fabrication is an inherent and foreseeable risk of the technology's architecture, thereby reinforcing, on scientific grounds, that the responsibility for the integrity of the judicial process cannot be delegated to the machine.

\section*{Acknowledgment}
We are extremely grateful to Daniela J. Restrepo and Jean Paul Roekaert for providing  us with a strong legal background to this issue of AI and law.

\appendix
\section*{Appendix}
This appendix supplies the step-by-step arithmetic underlying the tipping points described in the main text.
\section{Effective Attention Head Model}
\label{sec:basic-head}

We analyze a single self-attention head with greedy next-token selection (decoding temperature $T\rightarrow 0$). Each content type $X=\mathrm{A,B,C,D}$ is a vector $\vec{S}_X\in\mathbb{R}^d$. Let $z_1,\ldots,z_t$ denote the tokens seen so far (prompt + generated output). At position $t$, the attention score from the current query token $z_t$ to an earlier token $z_i$ is $s_{t,i}=\vec{S}_{z_t}\cdot\vec{S}_{z_i}$. The attention weight is
\[
a_{t,i} = \frac{\exp(s_{t,i}/T_{\mathrm{eff}})}{\sum_{j=1}^{t}\exp(s_{t,j}/T_{\mathrm{eff}})}\,,
\]
where $T_{\mathrm{eff}}$ is the effective temperature of the attention softmax (distinct from the decoding temperature $T$). The context vector is $\vec{N}(t)=\sum_{i=1}^{t}a_{t,i}\,\vec{S}_{z_i}$. Greedy decoding selects $z_{t+1}=\arg\max_x\,\vec{S}_x\cdot\vec{N}(t)$. We set the projection matrices $W_q=W_k=W_v=I$ for clarity; the tipping phenomenon persists for general $W_{q,k,v}$ but with less transparent algebra.

\subsection{Tipping-Point Formula}
For a prompt of $m$ $A$-tokens after which the model emits $n$ $B$-tokens before tipping to $D$, and with $\vec{S}_C$ perpendicular to the $A$--$B$--$D$ plane, the flip condition $\vec{S}_B\cdot\vec{N}=\vec{S}_D\cdot\vec{N}$ yields the tipping step
\begin{equation}
\label{eq:nstar}
n^{*}
=
\frac{m\,e^{\vec{S}_B\cdot\vec{S}_A/T_{\mathrm{eff}}}\;\vec{S}_A\!\cdot(\vec{S}_B-\vec{S}_D)}
     {e^{\vec{S}_B\cdot\vec{S}_B/T_{\mathrm{eff}}}\;\vec{S}_B\!\cdot(\vec{S}_D-\vec{S}_B)}\,.
\end{equation}
The ceiling $\lceil n^*\rceil$ predicts the length of the correct-output ($B$) block before a deterministic flip to harmful output ($D$). Equation~\eqref{eq:nstar} shows that tipping is governed by (i) prompt--$B$ alignment $\vec{S}_A\cdot\vec{S}_B$, (ii) the $B$--$D$ margin $\vec{S}_B\cdot(\vec{S}_D-\vec{S}_B)$, and (iii) the attention temperature $T_{\mathrm{eff}}$. Large $n^*$ implies a long run of seemingly correct output before sudden degradation---precisely the failure mode that eludes a lawyer's spot-check.

\section{Detailed Arithmetic for the Legal Brief Scenario}

The vectors are:
\begin{enumerate}
    \item $\vec{S}_A=(0.4,\,-0.3,\,0)$~~(Neutral factual basis)
    \item $\vec{S}_B=(0.8,\,\phantom{-}0.0,\,0)$~~(Correct legal application)
    \item $\vec{S}_C=(-0.2,\,-0.2,\,0)$~~(Anomalous legal query)
    \item $\vec{S}_D=(0.9,\,\phantom{-}0.5,\,0)$~~(Harmful legal falsehood)
\end{enumerate}

\subsection{First Tipping Point: Neutral to Correct}
The prompt is ACCA.

\textbf{Step 1: Token 5.}
Sequence: $\{A,C,C,A\}$. Context vector $\vec{N}=(0.140256,\,-0.256709,\,0.0)$.
\begin{align*}
    \vec{N}\cdot\vec{S}_A &= \mathbf{0.133115} \quad \textbf{(Max)} \\
    \vec{N}\cdot\vec{S}_B &= 0.112205
\end{align*}
Next token: A. History becomes ACCAA.

\textbf{Step 2: Token 6.}
Sequence: $\{A,C,C,A,A\}$. Context vector $\vec{N}=(0.197635,\,-0.266273,\,0.0)$.
\begin{align*}
    \vec{N}\cdot\vec{S}_A &= \mathbf{0.158936} \quad \textbf{(Max)} \\
    \vec{N}\cdot\vec{S}_B &= 0.158108
\end{align*}
Next token: A. History becomes ACCAAA.

\textbf{Step 3: Token 7 (first tip).}
Sequence: $\{A,C,C,A,A,A\}$. Context vector $\vec{N}=(0.234251,\,-0.272375,\,0.0)$.
\begin{align*}
    \vec{N}\cdot\vec{S}_A &= 0.175413 \\
    \vec{N}\cdot\vec{S}_B &= \mathbf{0.187401} \quad \textbf{(Max)}
\end{align*}
Selection rule flips. Next token: B. History becomes ACCAAAB.

\subsection{Second Tipping Point: Correct to Harmful}
The model emits $B$ tokens until the history reaches 12 tokens.

\textbf{Step 4: Token 13 (second tip).}
Sequence: $\{A,C,C,A,A,A,B,B,B,B,B,B\}$. Context vector $\vec{N}=(0.589815,\,-0.107221,\,0.0)$.
\begin{align*}
    \vec{N}\cdot\vec{S}_B &= 0.471852 \\
    \vec{N}\cdot\vec{S}_D &= \mathbf{0.477223} \quad \textbf{(Max)}
\end{align*}
Selection rule flips again. Next token: D. The second tipping point is reached, yielding the sequence $\{A,C,C,A,A,A,B,B,B,B,B,B,D,D,\ldots\}$ as shown in Fig.~1.


\end{document}